\documentclass[runningheads]{llncs}
\usepackage[T1]{fontenc}

\usepackage{graphicx}

\usepackage[hidelinks]{hyperref}

\usepackage{physics,amsmath,amssymb}

\usepackage[acronym]{glossaries}
\glsdisablehyper
\usepackage{siunitx}

\usepackage{booktabs}
\usepackage{xspace}
\usepackage{xcolor}
\usepackage{colortbl} 		
\usepackage{cleveref}
\usepackage{subcaption}
\usepackage{tikz}

\newcommand{\etal}{\textit{et al}.\@\xspace}

\newcommand{\wrt}{\textit{w}.\textit{r}.\textit{t}.\@\xspace}

\newcommand{\vc}[1]{\vb*{#1}}

\DeclareMathOperator{\argmin}{argmin}

\newlength{\defaulttabcolsep}
\newrobustcmd{\B}{\bfseries}

\newacronym{ad}{AD}{Anomaly Detection}
\newacronym{ae}{AE}{Autoencoder}
\newacronym{aevb}{AEVB}{Autoencoding Variational Bayes}
\newacronym{ann}{ANN}{Artificial Neural Network}
\newacronym{auroc}{AUROC}{Area Under the Receiver Operating Characteristic}
\newacronym{cdf}{CDF}{Cumulative Distribution Function}
\newacronym{dbs}{DBS}{Digitalisierung Bahnsystem}
\newacronym{elbo}{ELBO}{Evidence Lower Bound}
\newacronym{gan}{GAN}{Generative Adversarial Network}
\newacronym{gmm}{GMM}{Gaussian Mixture Model}
\newacronym{id}{ID}{In-Distribution}
\newacronym{iid}{i.i.d.\@}{independent and identically distributed}
\newacronym{kde}{KDE}{kernel density estimation}
\newacronym{kld}{KL divergence}{Kullback-Leibler divergence}
\newacronym{knn}{KNN}{k-Nearest Neighbor}
\newacronym{log}{LoG}{Laplacian of Gaussian}
\newacronym{lpips}{LPIPS}{Learned Perceptual Image Patch Similarity}
\newacronym{mse}{MSE}{mean squared error}
\newacronym{msp}{MSP}{Maximum Softmax Probability}
\newacronym{oe}{OE}{Outlier Exposure}
\newacronym{ood}{OOD}{Out-of-Distribution}
\newacronym{ood-od}{OOD-OD}{Out-of-Distribution Object Detection}
\newacronym{osr}{OSR}{Open-Set-Recognition}
\newacronym{pca}{PCA}{Principle Component Analysis}
\newacronym{pdf}{PDF}{Probability Density Function}
\newacronym{protovae}{ProtoVAE}{Prototypical Variational Autoencoder}
\newacronym{protodistvae}{ProtoDistVAE}{Prototypical Direct-Distance-Classifier VAE}
\newacronym{sem}{SEM}{self-explainable model}
\newacronym{sgd}{SGD}{Stochastic Gradient Descent}
\newacronym{sgvb}{SGVB}{Stochastic Gradient Variational Bayes}
\newacronym{ue}{UE}{Uncertainty Estimation}
\newacronym{umap}{UMAP}{Uniform Manifold Approximation and Projection}
\newacronym{vae}{VAE}{Variational Autoencoder}
\newacronym{xai}{XAI}{explainable AI}

\begin{document}

\title{Enclosing Prototypical Variational Autoencoder for Explainable Out-of-Distribution Detection}
\titlerunning{Enclosing ProtoVAE for explainable OOD Detection}
\author{Conrad Orglmeister\inst{1} \and
Erik Bochinski\inst{1} \and
Volker Eiselein\inst{1} \and
Elvira Fleig\inst{2}}
\authorrunning{C. Orglmeister et al.}
\institute{Digitale Schiene Deutschland, DB InfraGO AG, Berlin, Germany \and
\email{\{conrad.orglmeister,erik.bochinski,volker.eiselein\}@deutschebahn.com}\\
Communication Systems Group, Technische Universität Berlin, Berlin, Germany\\
\email{fleig@tu-berlin.de}}

\maketitle

\begin{abstract}
Understanding the decision-making and trusting the reliability of Deep Machine Learning Models is crucial for adopting such methods to safety-relevant applications.
We extend self-explainable Prototypical Variational models with autoencoder-based out-of-distribution (OOD) detection: A Variational Autoencoder is applied to learn a meaningful latent space which can be used for distance-based classification, likelihood estimation for OOD detection, and reconstruction.
The In-Distribution (ID) region is defined by a Gaussian mixture distribution with learned prototypes representing the center of each mode.
Furthermore, a novel restriction loss is introduced that promotes a compact ID region in the latent space without collapsing it into single points.
The reconstructive capabilities of the Autoencoder ensure the explainability of the prototypes and the ID region of the classifier, further aiding the discrimination of OOD samples. 
Extensive evaluations on common OOD detection benchmarks as well as a large-scale dataset from a real-world railway application demonstrate the usefulness of the approach, outperforming previous methods.

\end{abstract}

\begin{tikzpicture}[remember picture,overlay]
	\node[anchor=north,yshift=+100pt] at (current page.south) {
		\fbox{\parbox{1.2\textwidth}{
				\scriptsize This preprint has not undergone peer review or any post-submission improvements or corrections. The Version of Record of this contribution is published in \textbf{Computer Safety, Reliability and Security – SAFECOMP 2024 Workshops – DECSoS, SASSUR, TOASTS, and WAISE}, and is available online at \url{https://doi.org/10.1007/978-3-031-68738-9\_29}
		}}
	};
\end{tikzpicture}
\vspace{-30pt}

\section{Introduction}

Modern \glspl{ann} achieve remarkable results in recognizing patterns. However, due to their complexity and black-box character, their failures are hard to identify \cite{Lakshminarayanan2017} which limits their use in safety-critical environments. 
Additionally, certain common training schemes encourage overconfidence \cite{Guo2017}.
If \gls{ood} samples from other distributions than the \gls{id} training set are encountered in classification tasks, this issue persists.
Encountering such samples is often unavoidable in real-world applications, especially when operating in an open world as autonomous transportation systems do.
Therefore, \gls{ood} detection has arisen as the task of identifying instances not belonging to the training data distribution  
\cite{Yang2021a} which often means finding the label distribution but also
extends to identifying when the model might be unable to assess its input reliably. 
Anomaly detection, \gls{osr}, and Uncertainty Estimation are closely related to \gls{ood} detection and methods can often be applied to the other settings as well \cite{Yang2021a}. 
Most importantly, \gls{osr} requires explicitly classifying closed-world samples and detecting unknown classes from the open world \cite{Yang2021a}. 
 
Many \gls{ood} detection methods rely on post-hoc analysis of output or intermediate features from pre-trained classifiers but models trained solely for discrimination of \gls{id} categories may lack relevant features for OOD detection which limits the general usage of such approaches.
Integration of OOD detection into the classification framework is thus desirable, rather than applying it afterwards.

In this work, we extend the \gls{protovae} \cite{Gautam2022} to \gls{ood} detection.
Instead of the aforementioned post-analysis of application-specific pre-learned features for OOD detection, the feature space is designed to learn to distinguish unknown inputs from the beginning. 
This is done by estimating the training distribution, learning representations through reconstruction, and designing a distance-based latent space to quantify dissimilarity to \gls{id} clusters while also leveraging label information yielding promising results. Additionally, a restriction force is implemented to shape the latent \gls{id} region while reconstruction errors are used to identify remaining \gls{ood} samples mapped into this region as introduced in \cite{Zhou2022}. 

This work proposes the principle of an \emph{enclosing restriction} to decouple the previous trade-off between compression/estimation of the \gls{id} region and reconstructive quality to recover the input rather than just reconstruct features, thus alleviating \gls{ae}-based \gls{ood} detection by constraining the \gls{id} region in the latent space without collapsing it into one point.
To enhance the reconstructive power further, \gls{lpips} -- a perceptual metric -- is integrated into the framework for the reconstruction loss and \gls{ood} score. 
The generative and reconstructive abilities of the \gls{vae} framework enable the provision of additional information and explanation about extracted properties of the data distribution and certain samples, rendering the classification and OOD detection transparent.
The method is compared to state-of-the-art approaches using the OpenOOD \cite{Yang2022} and a custom railway benchmark.

\section{Related Work}

A \gls{protovae} architecture was presented by Gautam \etal \cite{Gautam2022} as a self-explainable model.  
Distance-based classification makes the decision more transparent and class distributions are divided into clusters. The ability to decode embeddings including prototypes fosters transparency \wrt learned data distribution. 
In this work, modifications enable more direct distance-based classification and enforce an \emph{enclosed} \gls{id} region making it ideal for \gls{ood} detection.

Yang \etal \cite{Yang2022} categorize \gls{ood} detection methods applied post-hoc, requiring training, Outlier Exposure, pre-processing, or data augmentation. 
Yang \etal \cite{Yang2021a} also distinguish approaches based on  
outputs of a classifier (classification-based), modeling the data distribution (density-based/generative), relying on distances in feature space (distance-based), and reconstructing the input measuring a reconstruction error (reconstruction-based). 
The approach of this work can be considered reconstruction-, distance- and density-based.
\gls{msp} as a baseline \gls{ood} score was examined by Hendrycks and Gimpel \cite{Hendrycks2017}. 
Hendrycks \etal \cite{Hendrycks2022a} use the maximum logit as a score (post-hoc). 
Sun \etal \cite{Sun2021} propose thresholding activations of the penultimate layer thus eliminating overconfidence caused by extreme activations.
Wang \etal \cite{Wang2022} design a virtual logit based on the smallest principle components.
Gal and Ghahramani \cite{Gal2016} apply \emph{Monte-Carlo dropout} during test time and 
Lakshminarayanan \etal \cite{Lakshminarayanan2017} train an ensemble of \glspl{ann}.
Hendrycks \etal \cite{Hendrycks2022b} propose a training-time augmentation based on fractals (\emph{PixMix}).

Nalisnick \etal \cite{Nalisnick2019} find that density estimates might assign higher likelihoods to \gls{ood} than to \gls{id} data. 
Xiao \etal \cite{Xiao2020} tackle this by retraining a \gls{vae}-encoder for a specific test-sample measuring a likelihood discrepancy. 
Sun \etal \cite{Sun2020} design  
a \gls{vae} with one Gaussian distribution per class. 
In contrast to this work, no perceptual metric, distance-based classification, or restriction-scheme for the \gls{id} region is used. 
Moreover, a custom probability is defined for a sample being part of a class distribution. There is a fixed threshold for the latter in contrast to the flexible \gls{ood} score fusion used in this work without a fixed threshold for one of the scores alone.
\emph{ARPL} \cite{Chen2022} generates near-\gls{ood} samples for learning adversarial reciprocal points representing individual negative classes.

Reconstructive \gls{ood} detection often involves elaborate schemes\cite{Denouden2018,Oza2019,Bercea2022,Zhou2022,Graham2023} as the reconstruction error alone often cannot separate \gls{ood} from \gls{id} data \cite{Denouden2018}. Existing approaches combine reconstruction error with Mahalanobis distance \cite{Denouden2018}, improve \gls{id} reconstruction with a deformation transformation\cite{Bercea2022} or use multiple reconstruction errors \cite{Oza2019,Graham2023}.
In \cite{Zhou2022}, the latent space region of an \gls{ae} to which \gls{id} samples are encoded (\emph{\gls{id} region}) is estimated by restricting \gls{id} data within the latent space. For \gls{ood} samples mapped into this region, the reconstruction error will be higher \cite{Zhou2022}. 
In contrast, in this work, an \emph{enclosing restriction} supports the trade-off between reliable estimation of the \gls{id} region and reconstruction quality.

Distance-based \gls{ood} detection involves Mahalanobis distance\cite{Lee2018} and \gls{knn} distance for pre-trained features. Requiring training, Deep SVDD \cite{Ruff2018} maps \gls{id} data into a hypersphere, and SIREN \cite{Du2022} discriminatively shapes representations using prototypes but not reconstruction.

\section{Methodology}

We introduce the \gls{protodistvae} for explainable \gls{ood} detection which extends the \gls{protovae} from \cite{Gautam2022} and further incorporates the principle of \gls{ae}-based \gls{ood} detection from \cite{Zhou2022}.
Following \cite{Zhou2022}, if an \gls{ae} reconstructs every \gls{id} sample sufficiently well and the \gls{id} region  $\mathcal{T}_\text{ID}$ can be estimated precisely, a sample can be concluded to be \gls{id} by fulfilling two conditions:
\begin{enumerate}
	\item An \gls{id} sample is embedded into $\mathcal{T}_\text{ID}$ (by definition).
	\item An \gls{id} sample exhibits a small reconstruction error.
\end{enumerate}
Under the given assumptions, \gls{ood} samples should never fulfill both conditions.

Our aim is to model a distribution of data that is representative for a set of prototypes. This means that different classes or parts of classes can be assigned to different sub-distributions during training, thus potentially increasing data diversity and simplifying the \gls{ood} detection. A distance metric space is learned where similar samples are in close proximity to each other.
\begin{figure}[t]
	\centering
	\includegraphics[width=0.9\linewidth]{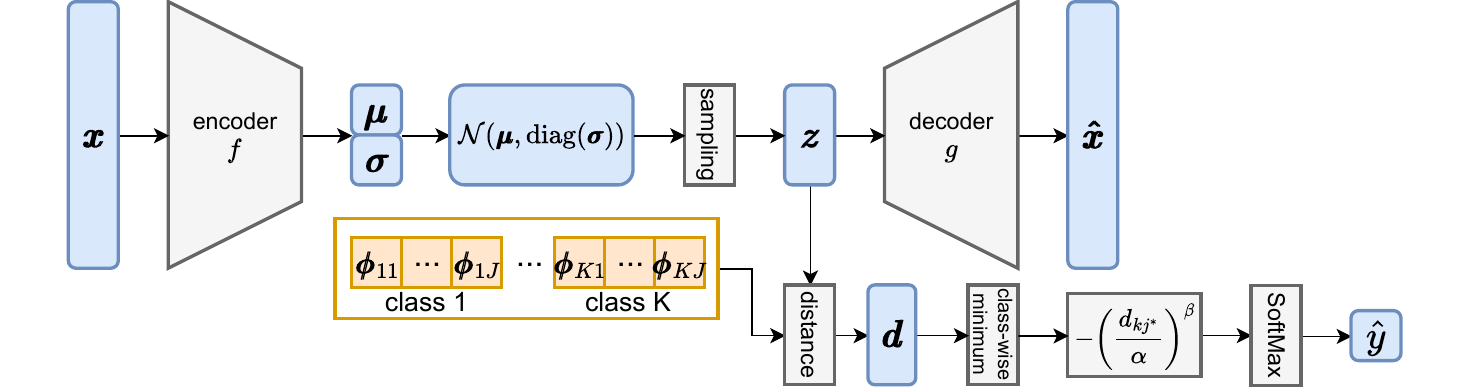} 
	\caption[Architecture of ProtoDistVAE]{ProtoDistVAE architecture: The input $\vc x$ is encoded into a latent Gaussian distribution from which a sample $\vc z$ is drawn and reconstructed to obtain $\vc{\hat{x}}$. Then, in the framework of generalized Gaussians, the SoftMax function returns the predicted probabilities and class estimate $\hat{y}$ for the distances to all prototypes.}
	\label{fig:protodistvae}
 \vspace{-0.4cm}
\end{figure}
Similar to \cite{Gautam2022}, we use an encoder $f_\psi$, a decoder $g_\theta$ and prototypes $\vc \phi_{kj}\in\mathbb{R}^L$ in an end-to-end trainable fashion (see \Cref{fig:protodistvae}).
The rows of matrix $\vc \Phi_k \in \mathbb{R}^{J\cross L}$ describe the $J$ prototype vectors of class $k \in K$ classes.


Given a training dataset $\mathcal{D}=\{(\vc{x}^1, (\vc{x}^1,y^1)), ..., (\vc{x}^N, (\vc{x}^N,y^N))\}$ with $N$ labeled samples, the input $\vc x^i$ itself yields the target variables for reconstruction and a class label $y^i$.
The model is trained as a \gls{vae} learning a Gaussian mixture distribution where the encoder embeds the input $\vc x^i$ to a posterior Gaussian distribution $p(\vc{z}|\vc{x}^i) = \mathcal{N}(\vc{z}; \vc{\mu}^i, \text{diag}((\vc{\sigma}^i)^2))$ in the latent domain.  
During training, a latent representation $\vc{z}^i$ is sampled whereas during inference, the mean value is used for the latent representation the decoder processes into the image space reconstruction $\hat{\vc{x}}^i$.

For classification, the Euclidean distances of the latent variable to all prototypes are computed (\Cref{eq:protovae_distances}) and the minimum distance of each class yields the closest prototype. 
It is important to minimize the distance of an embedding to only one prototype distribution during training.
The distances are transformed into logits by the generalized Gaussian distribution for enclosing restriction and are fed into a SoftMax function to obtain a purely distance-based, latent space classification without a learnable classifier.
\begin{align}
	d(\vc z^i, \vc{\phi}_{kj}) &= d^i_{kj} = \| \vc z^i - \vc{\phi}_{kj} \|_2  \label{eq:protovae_distances} \\
        P_\psi(y = k | \vc{x}^i) &= \frac{\exp(l_k^i)}
	{\sum_{k'=1}^{K} \exp(l_{k'}^i)} \ , \quad 
	l_{k'}^i = -\left(\frac{d_{k'j^*(k')}}{\alpha}\right)^\beta 
	\label{eq:protodistvae_classification} \\
	j^*(k) &= \argmin_j(d_{kj})
\end{align}
The original \gls{protovae} architecture uses a linear classifier and distance-based similarity scores
\cite{Gautam2022}. Similarity scores exhibit large gradients for embeddings close to a prototype which potentially leads to embeddings collapsing into the respective prototype position, and thus to degradation of reconstruction quality when different embeddings are not encoded differently. 
As a remedy, \gls{protodistvae} uses an enclosing restriction leading to weaker gradients close to prototypes. Embeddings shall be trapped in a certain \emph{\gls{id} region}, but inside, the coding of embeddings shall be unconstrained.
For this reason, generalized Gaussian distributions are used in the classification layer where $\alpha$ defines the width of the distribution and $\beta \ge 2$ controls the shape and "enclosedness" of the distribution.

In order to not distort the distance metric space, \gls{protodistvae} uses distances more explicitly for classification. The linear classifier which essentially calculates a \emph{sum} of distances is replaced by
using only the minimum distances to prototypes per class. These are translated into logits $l^i_{k'}$ by the framework of generalized Gaussians and probabilities using the SoftMax function (\Cref{eq:protodistvae_classification}). 
Cross-entropy is then applied to the modified predicted probabilities. $j^*(k)$ is the nearest prototype within class $k$ while $\vc{d^*}$ is the minimum distances vector for every class.
Thus, instead of a sum of distances to multiple prototypes, the distance to only one prototype is minimized for a specific embedding.

The overall loss consists of a sum of four terms: The cross-entropy loss $\mathcal{L}'_{\text{cls}}$ shown in \Cref{eq:protovae_loss_ce} provides label information to enable the network to extract useful embeddings for discrimination and minimize the embedding distance to prototypes of the correct class. 
Each class is modeled by a mixture of $J$ normal distributions centered around the respective class prototypes for \gls{vae}-like distribution estimation and \gls{kld} \wrt the nearest prototype distribution of the correct class is computed to obtain the loss $\mathcal{L}'_{\text{KL}}$ (\Cref{eq:protodistvae_loss_kl}).
The reconstruction loss aims to recover the input samples \cite{Gautam2022} by separating groups of samples near each other for a better reconstruction. We use the \gls{lpips} metric \cite{Zhang2018} for this task as it gives a more robust similarity between images than traditional metrics as e.g. \gls{mse} by using a calibrated pre-trained network aligned towards human perception \cite{Zhang2018}. 

In order to prevent the collapse of prototypes of a class, an orthonormalization loss $\mathcal{L}_\text{orth}$ (\Cref{eq:protovae_loss_orth}) is used to encourage prototypes within a class (after subtracting their mean $\bar{\vc{\phi}}_k$) to be orthonormal to each other \cite{Gautam2022}. 
It is defined as the average of the class-wise Frobenius norms $\|\cdot\|_F$.
\begin{align}
	\mathcal{L}'_{\text{cls}}(\vc\psi, \vc \Phi; \vc x^i, k) &= -\log P_\psi(y = k| \vc{x}^i) \label{eq:protovae_loss_ce} \\
    \mathcal{L}'_{\text{KL}}(\vc\psi, \vc\Phi_k; \vc x^i, k\!=\!y^i) &=    
	D_{KL}( \mathcal{N}(\vc{\mu}^i, \text{diag}((\vc{\sigma}^i)^2))\| \mathcal{N}(\vc\phi_{kj^*(k)}, \vc I_L) ) \label{eq:protodistvae_loss_kl} \\ 
    \mathcal{L}'_{\text{rec}}(\vc\psi, \vc \theta; \vc{x}^i, \vc{\hat{x}}^i) &= e_\text{LPIPS}( \vc{x}^i, \vc{\hat{x}}^i) \label{eq:protovae_loss_lpips} \\
	\mathcal{L}_{\text{orth}}(\vc\Phi) &= \frac{1}{K}\sum\nolimits_{k=1}^{K} \|\tilde{\vc\Phi}_k \tilde{\vc\Phi}_k^T - \vc I_J\|_F^2 , \  \tilde{\vc\Phi}_k\!=\!(\vc\phi_{kj} - \bar{\vc\phi}_k)_{j=1..J} \label{eq:protovae_loss_orth}
\end{align}
In summary, \gls{protodistvae} introduces LPIPS \cite{Zhang2018} as reconstruction loss and replaces the linear classifier layer as well as similarity scores by direct minimum distances and the framework of generalized Gaussians to implement an enclosing restriction loss. The complete loss function is:
\begin{align}
	\mathcal{L} = 
	w_\text{cls} \mathcal{L}_{\text{cls}} + 
	w_\text{KL} \mathcal{L}_{\text{KL}} +
	w_\text{rec} \mathcal{L}_{\text{rec}} +
	w_\text{orth} \mathcal{L}_{\text{orth}}
\end{align}

For OOD detection, a distance-based OOD score 
and the \gls{lpips} reconstruction error 
are merged. 
During experimentation, we found that the minimum distance to the next prototype can be improved by using the \gls{msp} score $\lambda_\text{MSP}\!=\!\max_k P_\psi(y=k|\vc x^i)$ in the \gls{protodistvae} context which is the probability that an embedding belongs to the most likely generalized Gaussian under condition 
that it is \gls{id}. 
As \gls{protodistvae} relies on distances for classification, \gls{msp} is also distance-based.  
Also the $\lambda_\text{DistRatio} = \sum_j d_{\widehat{k}j}/(\sum_{k}\sum_{j} d_{kj})$ is applied where $\widehat{k}$ indicates the predicted class. We assume these scores perform better than the minimum distance because the class distribution in the latent space might be skewed and \gls{ood} samples are embedded between different class regions.

For fusion of scores, one distance score and one reconstruction error are normalized \wrt to their validation set distributions to make them comparable using a lower and upper percentile of the score distribution to obtain the normalized score $\widetilde{\lambda}(\vc x) = (\lambda(\vc x) - \lambda_\text{lower}) / (\lambda_\text{upper} - \lambda_\text{lower})$. 
Both score types are combined into one score using $L_2$ or $L_\infty$ norm: $\lambda_{L_p}(\vc x) = \|(  \widetilde{\lambda}_1(\vc x), \widetilde{\lambda}_2(\vc x) )^T \|_p$ where $p$ denominates the degree. The $L_\infty$ norm tends to reflect a hard decision (e.g. at least one score is above its threshold) and the $L_2$ norm a flexible decision (one score is too high or both together are rather high and therefore indicates an OOD sample). This type of fusion means that no probabilities need to be modeled explicitly and thus avoids any need for modeling assumptions.

\section{Experimental Results}

For numerical evaluation, we compare our approach to the state-of-the-art based on the OpenOOD benchmark \cite{Yang2022} and a non-public dataset from the railway domain (DBS dataset).
A general advantage of the proposed method is that it allows human insights into the training distribution and decision-making of the network
by reconstructing samples, prototypes, and distances in the latent space which 
supports its usage in safety-critical domains.

\paragraph{General Experimental Setup}
The OpenOOD benchmark provides implementations of state-of-the-art approaches for comparison and defines sub-benchmarks according to the ID datasets MNIST, CIFAR10, CIFAR100, and ImageNet. Another dataset is then used as OOD data. Datasets are labeled as near OOD or far OOD according to their ID similarity, e.g. if they have similar color distributions. Open Set Recognition (OSR) is also provided by partitioning a dataset into ID and OOD classes. M-6 benchmark is based on MNIST, C-6 on CIFAR-10, C-50 on CIFAR-100, and T-20 on TinyImageNet with the numeral representing the number of ID classes.

The DBS dataset was collected from video recordings of a camera mounted on a commuter train in a typical operation. 
Object proposals were automatically collected and classified into trains and persons. 
The annotations were manually checked and \gls{ood} samples (i.e. false positive detections) were placed in a separate category.  
In our evaluation, we used $8351$ samples of people, $8340$ samples of trains, and $5001$ non-objects labeled as OOD, all rescaled to size $64 \times 64$.
Person and train samples were divided equally into training (60\%), validation (10\%), and test (30\%) splits (OOD samples used only for testing). 
We use $J\!=\!1$ prototype per class in all experiments as a higher number did not improve the performance.

The generalized Gaussian parameters $\alpha$ and $\beta$ were both set to $2$ for all experiments.
The encoder was chosen as ResNet-50 \cite{He2016} for ImageNet and as ResNet-18 for all benchmarks with $64\times64$ sized images (including the DBS dataset) and $32\times32$ sized images. A convolutional encoder with five layers was used for all $28\times28$ sized images, for the decoder a five-layered network using subpixel-convolutions \cite{Shi2016} is used. For ImageNet the decoder consists of seven layers and for all other benchmarks, it consists of six layers. The latent dimensionality $L$ is chosen as $1/3$, $1/24$ or $1/96$ of the input dimensionality. 
After training, \gls{id} validation data were used for normalization of the \gls{ood} scores, which are used afterwards for score fusion during testing. For evaluation, \gls{id} classification performance is measured in accuracy and \gls{ood} detection performance in \gls{auroc}. \gls{auroc} is a threshold-independent metric and measures how well a score separates \gls{id} and \gls{ood}.

\subsection{OOD Detection Performance}
\Cref{tab:sota} shows the \gls{ood} detection performance in terms of AUROC compared to state-of-the-art methods. 
\begin{table}[t]
	\tiny	
	\centering
	\setlength{\tabcolsep}{2pt}
	\caption{OOD detection performance (AUROC in \%) on OpenOOD benchmark and CIFAR-100 ID accuracy (\%) for different approaches: Best performances marked in bold. Results from other methods taken from \cite{Yang2022}. }
	\begin{tabular}{@{}p{2.0cm}S[,table-format=3.1]S[,table-format=3.1]S[,table-format=3.1]S[,table-format=3.1]S[,table-format=3.1]S[,table-format=3.1]S[,table-format=3.1]S[,table-format=3.1]S[,table-format=3.1]S[,table-format=3.1]S[,table-format=3.1]S[,table-format=3.1]S[,table-format=3.1]@{}}
		\toprule
		Method                                   & {M-6}         & {C-6}         & {C-50}        & {T-20}        & \multicolumn{2}{c}{MNIST}      & \multicolumn{2}{c}{CIFAR-10}  & \multicolumn{2}{c}{CIFAR-100} & \multicolumn{2}{c}{ImageNet}  & Acc           \\
		& {osr}         & {osr}         & {osr}         & {osr}         & {near}        & {far}          & {near}        & {far}         & {near}        & {far}         & {near}        & {far}         & {CIFAR-100}   \\ 
		\midrule
            ARPL \cite{Chen2022}                & {-}          & {-}          & {-}          & {-}          & 93.9       & 99.0         & 87.2         & 88.0         & 74.9         & 74.0          & {-}           & {-}           & 71.7           \\
		Mahalanobis \cite{Lee2018}               & 89.8          & 42.9          & 55.1          & 57.6          & 98.0          & 98.1           & 66.5          & 88.8          & 51.4          & 70.1          & 68.3          & 94.0          & 75.8          \\
		KNN \cite{Sun2022}                       & 97.5          & 86.9          & \textbf{83.4} & 74.1          & 96.5          & 96.7           & 90.5          & 92.8          & 79.9          & 82.2          & \textbf{80.8} & \textbf{98.0} & 77.1          \\
		ReAct \cite{Sun2021}                     & 82.9          & 85.9          & 80.5          & 74.6          & 90.3          & 97.4           & 87.6          & 89.0          & 79.5          & 80.5          & 79.3          & 95.2          & 75.8          \\
		MaxLogit \cite{Hendrycks2022a}           & 98.0          & 84.8          & 82.7          & 75.5          & 92.5          & 99.1           & 86.1          & 88.8          & 81.0          & 78.6          & 73.6          & 92.3          & 77.1          \\
		Dropout \cite{Gal2016}                   & 96.2          & 84.5          & 81.1          & 73.6          & 91.5          & 97.1           & 87.3          & 90.4          & 80.1          & 79.4          & {-}           & {-}           & 77.1          \\
		DeepEnsemble \cite{Lakshminarayanan2017} & 97.2          & 87.8          & 83.1          & \textbf{76.0} & 96.1          & 99.4           & 90.6          & 93.2          & \textbf{82.7} & 80.7          & {-}           & {-}           & \textbf{80.5} \\
		PixMix \cite{Hendrycks2022b}             & 93.9          & \textbf{90.9} & 78.0          & 73.5          & 93.7          & 99.5           & \textbf{93.1} & \textbf{95.7} & 79.6          & \textbf{85.5} & {-}           & {-}           & 77.1          \\
		ProtoDistVAE                             & \textbf{98.4} & 76.6          & 69.0          & 62.4          & \textbf{99.9} & \textbf{100.0} & 80.0          & 90.6          & 65.3          & 74.2          & 69.6          & 80.1          & 48.8         \\
		\bottomrule
	\end{tabular}
	\label{tab:sota}
        \vspace{-0.4cm}
\end{table}
\gls{protodistvae} was trained using only \gls{lpips} reconstruction loss with weight $w_\text{rec}=1$. Cross-entropy and \gls{kld} loss were used similarly with a weight of $w_\text{cls}=w_\text{KL}=1$. Distance ratio $\lambda_\text{DistRatio}$ and \gls{lpips} $\lambda_\text{LPIPS}$ were used as scores to be fused by $L_\infty$ norm. The latent space dimensionality $L$ was chosen as $1/24$ of the input dimensionality.

Compared to the other methods, ProtoDistVAE performs best on the MNIST-based benchmarks. This is likely due to its low diversity, making it easier to learn a latent distribution.
For CIFAR10, ProtoDistVAE performs on par with other methods.
However, the performance for highly diverse datasets with a large number of classes decreases as ID estimation and classification are performed in the same latent space and may impair each other.
Similarly, higher resolutions lead to difficulties for ProtoDistVAE in detecting OOD samples, likely due to the increased complexity of reconstruction.

\begin{figure}[t]
	\centering
	\begin{subfigure}{0.32\textwidth}
		\centering
		\includegraphics[width=1\linewidth,trim=1cm 0 0.8cm 0]{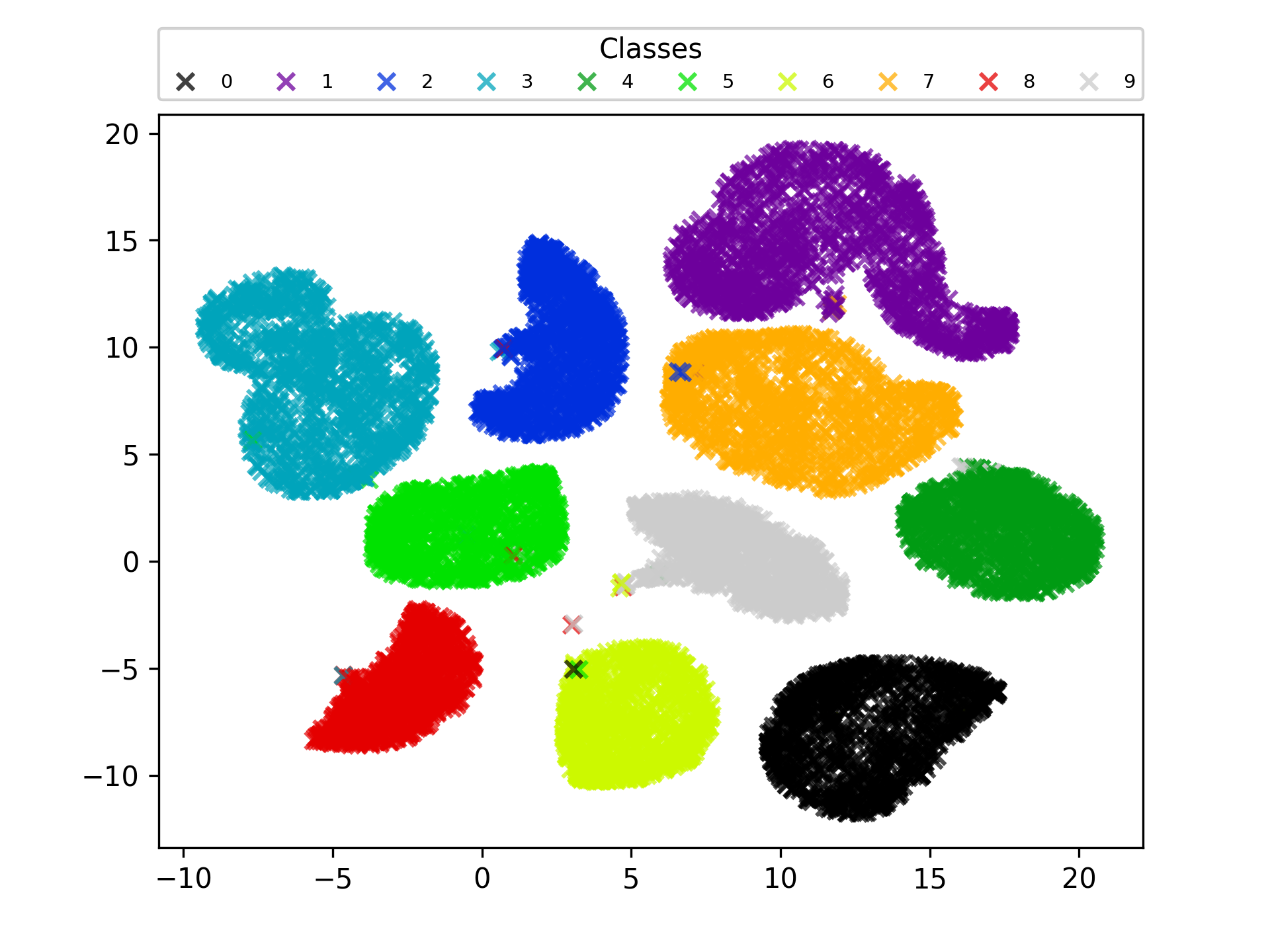}
		\caption{MNIST}
		\label{fig:umap-mnist}
	\end{subfigure}
	\begin{subfigure}{0.32\textwidth}
		\centering
		\includegraphics[width=1\linewidth,trim=1cm 0 0.8cm 0]{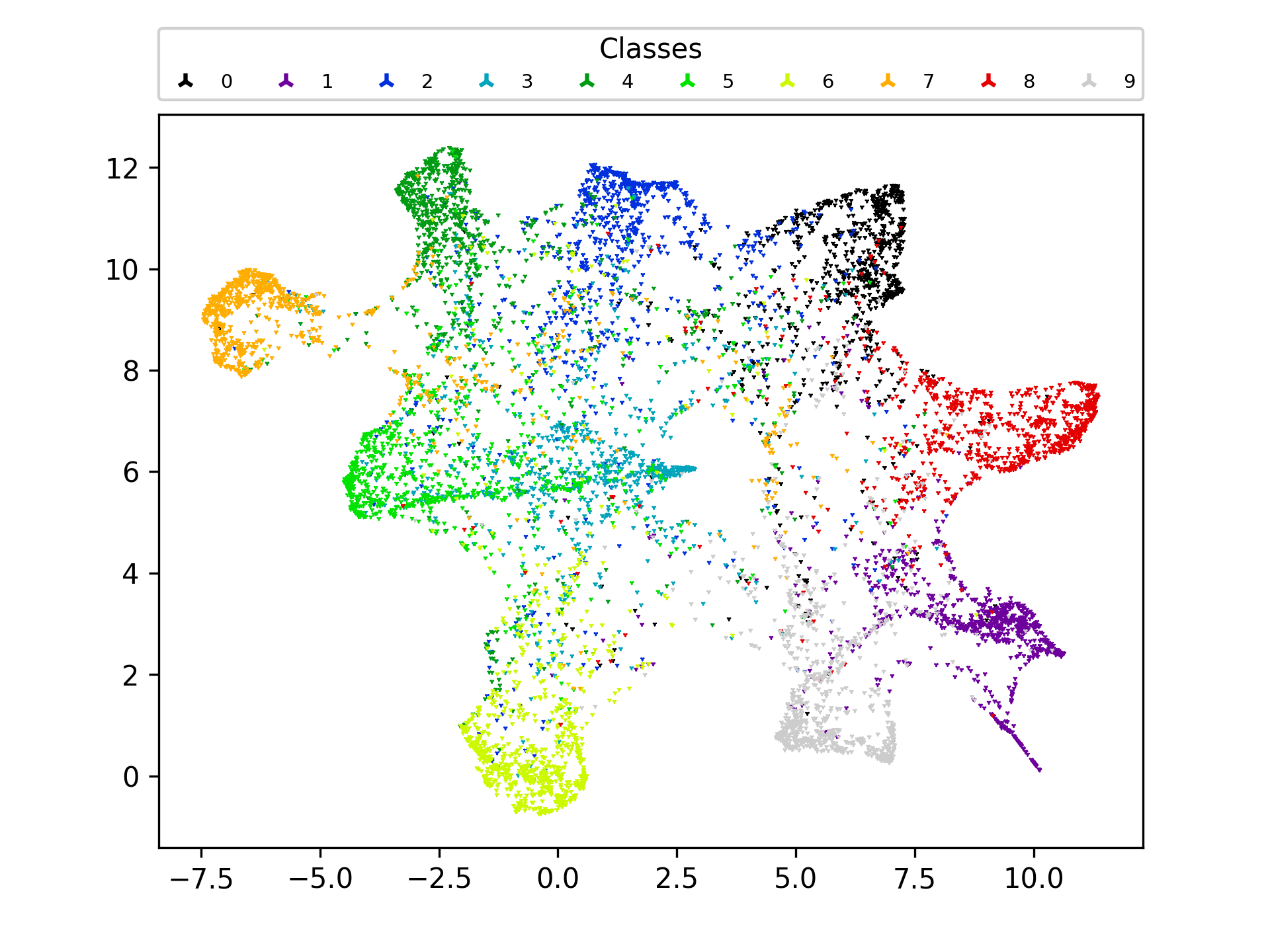}
		\caption{CIFAR10}
		\label{fig:umap-cifar}
	\end{subfigure}
	\begin{subfigure}{0.32\textwidth}
		\centering
		\includegraphics[width=1\linewidth,trim=1cm 0 0.8cm 0]{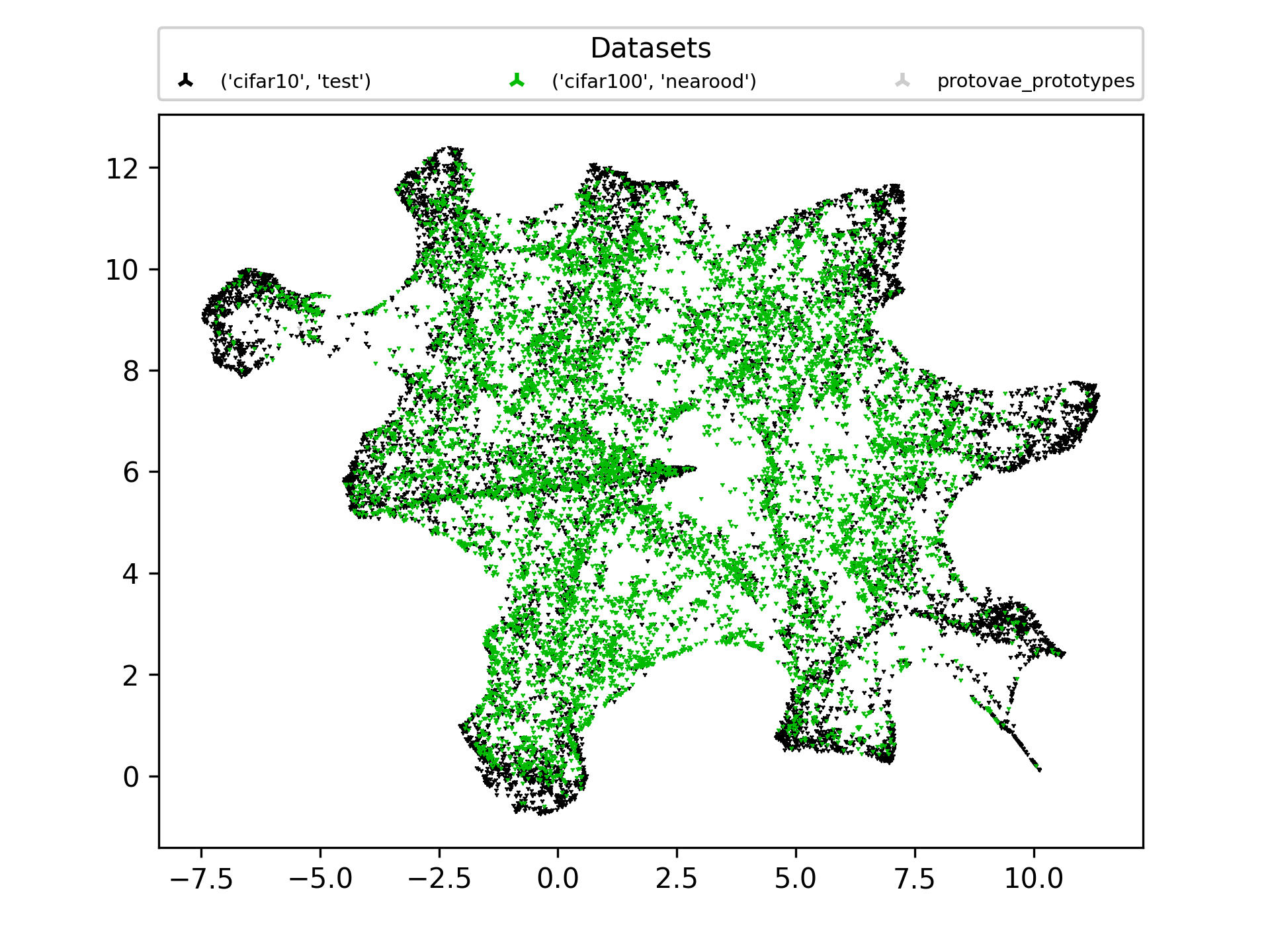}
		\caption{CIFAR10 -- CIFAR100}
		\label{fig:umap-cifar-id-ood}
	\end{subfigure}
 \vspace{-0.2cm}
	\caption{\gls{umap} visualization of the latent space embeddings of trained \glspl{protodistvae}. (\subref{fig:umap-mnist}) On MNIST, color-coded classes are clearly separated. (\subref{fig:umap-cifar}) On CIFAR10, clusters blend into each other. (\subref{fig:umap-cifar-id-ood}) ID (CIFAR10) versus OOD (CIFAR100): embedding of OOD samples appears mainly between class prototypes.}
	\label{fig:umap-mnist-cifar}
    \vspace{-0.4cm}
\end{figure}

\Cref{fig:umap-mnist-cifar} shows further insights through an \gls{umap} visualization of the latent space and illustrates how our method allows understanding its decision-making. The method works best in cases of clearly separable datasets and performs worse if data cannot be attributed well to clusters extracted. However, it should be mentioned that CIFAR10 vs. CIFAR100 is generally a hard OOD benchmark. 
\gls{id} samples in the space between prototypes might be interesting for further analysis since these exhibit a higher uncertainty and could be exploited by active learning or for identifying a sample with very different attributes within a class.

\Cref{tab:sota-bahn} shows some results on the DBS dataset. Here, an increased 
weight on \gls{lpips} ($w_\text{rec}=100$) was used to improve the \gls{ood} detection performance without harming classification accuracy. 
\begin{table}[t]
    \caption{Experimental results of \gls{ood} detection in AUROC (\%) and \gls{id} accuracy (\%): (a) DBS dataset results of state-of-the-art methods (parameterized as in \cite{Yang2022}) compared to ProtoDistVAE with LPIPS score combined by $L_\infty$ fusion with DistRatio and MSP, respectively. (b) ProtoVAE vs. ProtoDistVAE. (c) Influence of reconstruction loss when using \gls{lpips} as \gls{ood} score. }
    \begin{subtable}{0.29\textwidth}
        \centering
        \tiny
        \def\arraystretch{1.0}
        \caption{DBS dataset
        }
        \begin{tabular}{@{}lS[,table-format=2.1]S[,table-format=2.1]@{}}
            \toprule
            Method                               & AUC         & Acc           \\ \midrule
            ARPL \cite{Chen2022}                 & 80.2          & \textbf{99.7} \\ 
            Mahalanobis \cite{Lee2018}           & 75.4          & 93.6          \\
            KNN \cite{Sun2022}                   & 84.9          & 99.6          \\
            MSP \cite{Hendrycks2017}             & 83.6          & 99.6          \\
            MaxLogit \cite{Hendrycks2022a}       & 83.6          & 99.6          \\
            ReAct \cite{Sun2021}                 & 81.8          & 99.6          \\
            VIM \cite{Wang2022}                  & 85.4          & 99.6          \\
            Dropout \cite{Gal2016}               & 79.8          & \textbf{99.7} \\
            DeepEnsemble \cite{Lakshminarayanan2017} & 83.4      & \textbf{99.7} \\
            Pixmix \cite{Hendrycks2022b}         & \textbf{89.3} & 99.6          \\        \midrule
            LPIPS+DistRatio                      & 87.9          & 99.5          \\
            LPIPS+MSP                            & 88.8          & 99.5          \\ 
                \bottomrule
        \end{tabular}
        \label{tab:sota-bahn}
    \end{subtable}
    \begin{subtable}{0.7\textwidth}
        \tiny
	\centering
	\setlength{\tabcolsep}{1pt}
        \setcounter{subtable}{2}
	\caption{OpenOOD benchmark (partial): reconstruction loss}
	\begin{tabular}{>{\columncolor[gray]{0.9}}c>{\columncolor[gray]{0.9}}cS[detect-weight,table-format=3.1]S[detect-weight,table-format=3.1]S[detect-weight,table-format=3.1]S[detect-weight,table-format=3.1]S[detect-weight,table-format=3.1]S[detect-weight,table-format=3.1]S[detect-weight,table-format=3.1]S[detect-weight,table-format=3.1]|S[detect-weight,table-format=2.1]S[detect-weight,table-format=2.1]}
\toprule
{$L$} & {Loss} &  \multicolumn{2}{c}{mnist} & \multicolumn{2}{c}{cifar10} & \multicolumn{2}{c}{cifar100} & \multicolumn{2}{c}{ImageNet} & {cifar10} & {cifar100} \\
{} & {} & {near} & {far} & {near} & {far} & {near} & {far} & {near} & {far} & {ID} & {ID} \\
{} & {} & {AUC} & {AUC} & {AUC} & {AUC} & {AUC} & {AUC} & {AUC} & {AUC} & {Acc} & {Acc} \\
\midrule
1/24 & MSE  & 99.4 & \B 100.0 & 62.5 & 44.5 & 58.8 & 47.6 & 57.1 & 33.8 & 79.4 &  46.7 \\
1/96 & MSE  & 99.4 & \B 100.0 & 63.7 & 46.8 & 59.2 & 48.9 & 52.4 & 34.0 & 78.7 & 46.5 \\
\midrule
1/24 & LPIPS  & \B 100.0 & \B 100.0 & \B 78.3 & \B 90.8 & \B 61.0 & 71.4 & 56.6 & 37.5 & 79.9 & \B 48.8 \\
1/96 & LPIPS  & \B 100.0 & \B 100.0 & 77.7 & 90.1 & 60.8 & \B 80.6 & \B 60.2 & 48.4 & \B 80.3 & 46.5 \\
\bottomrule
\end{tabular}

	\label{tab:lpips-score}
    \end{subtable}
    \begin{subtable}{1\textwidth}
	\tiny
	\centering
	\setlength{\tabcolsep}{0.7pt}
        \setcounter{subtable}{1}
	\caption{OpenOOD benchmark: ProtoVAE vs. ProtoDistVAE using MSP score}
	\begin{tabular}{>{\columncolor[gray]{0.9}}c>{\columncolor[gray]{0.9}}cS[detect-weight,table-format=3.1]S[detect-weight,table-format=3.1]S[detect-weight,table-format=3.1]S[detect-weight,table-format=3.1]|S[detect-weight,table-format=3.1]S[detect-weight,table-format=3.1]S[detect-weight,table-format=3.1]S[detect-weight,table-format=3.1]S[detect-weight,table-format=3.1]S[detect-weight,table-format=3.1]S[detect-weight,table-format=3.1]S[detect-weight,table-format=3.1]|S[detect-weight,table-format=2.1]S[detect-weight,table-format=2.1]}
\toprule
{arch} & {$L$} & \multicolumn{1}{c}{mnist6} & \multicolumn{1}{c}{cifar6} & \multicolumn{1}{c}{cifar50} & \multicolumn{1}{c}{tin20} & \multicolumn{2}{c}{mnist} & \multicolumn{2}{c}{cifar10} & \multicolumn{2}{c}{cifar100} & \multicolumn{2}{c}{ImageNet} & {cifar10} & {cifar100} \\
{} & {} & {osr} & {osr} & {osr} & {osr} & {near} & {far} & {near} & {far} & {near} & {far} & {near} & {far} & {ID} & {ID} \\
{} & {} & {AUC} & {AUC} & {AUC} & {AUC} & {AUC} & {AUC} & {AUC} & {AUC} & {AUC} & {AUC} & {AUC} & {AUC} & {Acc} & {Acc} \\
\midrule
ProtoVAE & 1/3 & 94.1 & 66.1 & 65.3 & 62.4 & 92.1 & 98.6 & 71.2 & 75.9 & 61.5 & 57.3 & 55.6 & 30.7 & 76.1 & 39.8 \\
ProtoVAE & 1/24 & 95.7 & 68.6 & 66.2 & 61.4 & 92.3 & \B 99.1 & 71.7 & 75.1 & 62.9 & 56.9 & 49.9 & 49.9 & 80.1 & 43.0 \\
ProtoVAE & 1/96 & 96.5 & 65.4 & 64.8 & 62.5 & 94.9 & 98.4 & 71.5 & 76.8 & 62.5 & 61.5 & 50.1 & 50.1 & 79.4 & 41.0 \\
ProtoDistVAE$\ $ & 1/3 & \B 97.9 & 76.5 & 63.7 & 60.5 & \B 96.7 & 99.0 & 75.6 & 73.9 & 63.9 & 51.1 & 60.7 & 70.7 & \B 81.3 & 41.0 \\
ProtoDistVAE$\ $ & 1/24 & 96.4 & 76.0 & \B 67.4 & 59.4 & 91.4 & 93.4 & \B 76.3 & 76.3 & 64.5 & \B 62.2 & \B 67.3 & \B 78.6 & 79.4 & \B 46.7 \\
ProtoDistVAE$\ $ & 1/96 & 96.1 & \B 77.3 & 66.8 & \B 63.1 & 94.5 & 96.5 & 75.4 & \B 76.9 & \B 65.1 & 56.4 & 50.1 & 50.1 & 78.7 & 46.5 \\
\bottomrule
\end{tabular}

	\label{tab:15-msp}
    \end{subtable}
    \vspace{-.3cm}
\end{table}
The accuracy is on par with other methods, likely due to only two classes being available. For OOD detection, PixMix and ProtoDistVAE perform best, while VIM and KNN also show good results. Combining $\lambda_\text{LPIPS}$ with $\lambda_\text{MSP}$ 
further improves the results with a gain of $0.9\%$.

ProtoDistVAE performs well on the DBS dataset due to its composition. The data samples are often quite similar as trains and persons are captured from the same angles and there are little variations e.g. in perspective, weather, lighting, and color. In comparison, ImageNet shows more inconsistent data with more diverse appearances across the same class. ProtoDistVAE benefits from a reduced intra-class variance and “complete” data distribution which allows it to model the data more easily. Hypothetically, it is easier for the network to recognize systematics in the data.
PixMix augmentation seems to benefit from a complete distribution and even further increases the diversity of the data. However, the data distribution is not represented in the model and classification is not transparent.
Other methods perform worse: Ensembling shows a lower-than-usual performance as it depends on variations in the prediction of individual networks and these variations are weaker due to low data diversity in this dataset. Methods depending purely on classification-based schemes might suffer from overconfidence due to easier classification across only two classes and low data diversity. ProtoDistVAE, however, does not overfit for classification and aims to learn a representation of the data. In addition, the reconstruction error helps it to identify overconfidently classified samples mapped into its ID-region.

\subsection{Ablation Study: ProtoVAE vs. ProtoDistVAE} 

Comparing the proposed \gls{protodistvae} architecture to the base \gls{protovae}
, the reconstruction loss was set to a constant level. This does not change reconstruction error-based OOD detection according to the observed data. \Cref{tab:15-msp} shows detection results for ProtoVAE and ProtoDistVAE using the distance-based MSP score based on the predicted probabilities. Note that an improved distance-based score potentially increases performance even further when fused with a reconstruction error score.
ProtoDistVAE outperforms ProtoVAE in almost all benchmarks for OOD detection and for different values of the latent dimension $L$ which can be explained by the direct use of distances for classification and the enclosing restriction used during training. 
The latter actively shapes the ID-region by trapping the ID embeddings in the proximity of the class-specific prototypes.
Furthermore, the results display the importance of the latent dimensionality $L$ for both networks. Different values for $L$ are optimal for different levels of complexity reflected in different datasets. Too low values reduce the information coded in the representation while too high values inhibit a clear assignment of samples to class prototypes.

\subsection{Reconstruction}

\begin{figure}[t]
	\centering
	\begin{subfigure}{0.47\textwidth}
		\centering
            \includegraphics[width=1\linewidth]{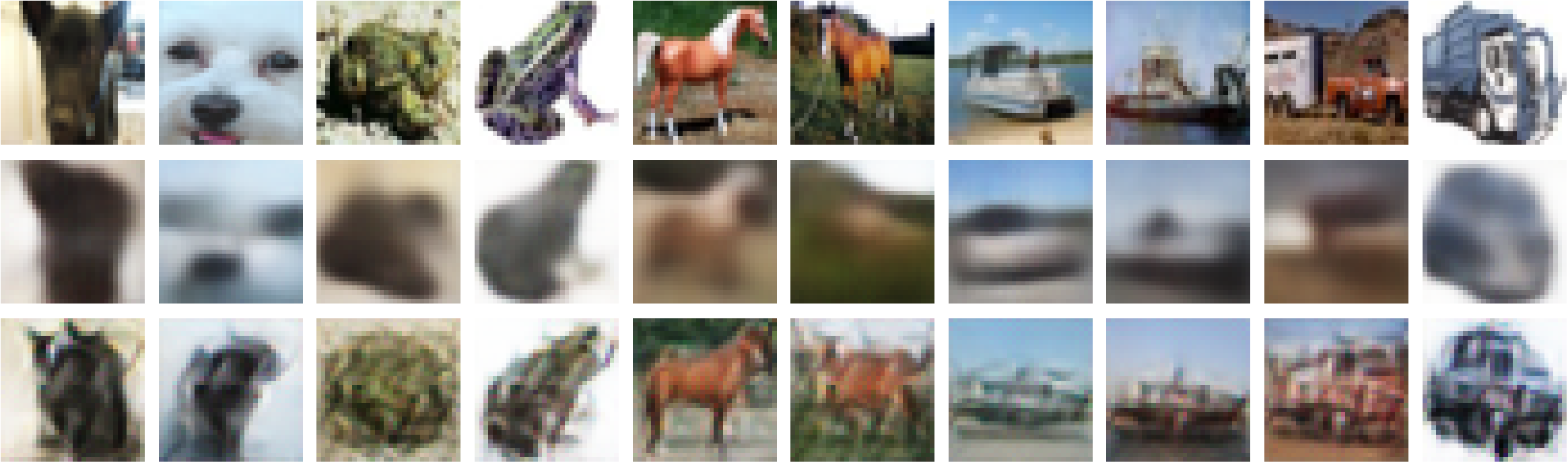}
		\caption{CIFAR10 (ID)}
		\label{fig:cifar10-id}
	\end{subfigure}
	\begin{subfigure}{0.47\textwidth}
		\centering
            \includegraphics[width=1\linewidth]{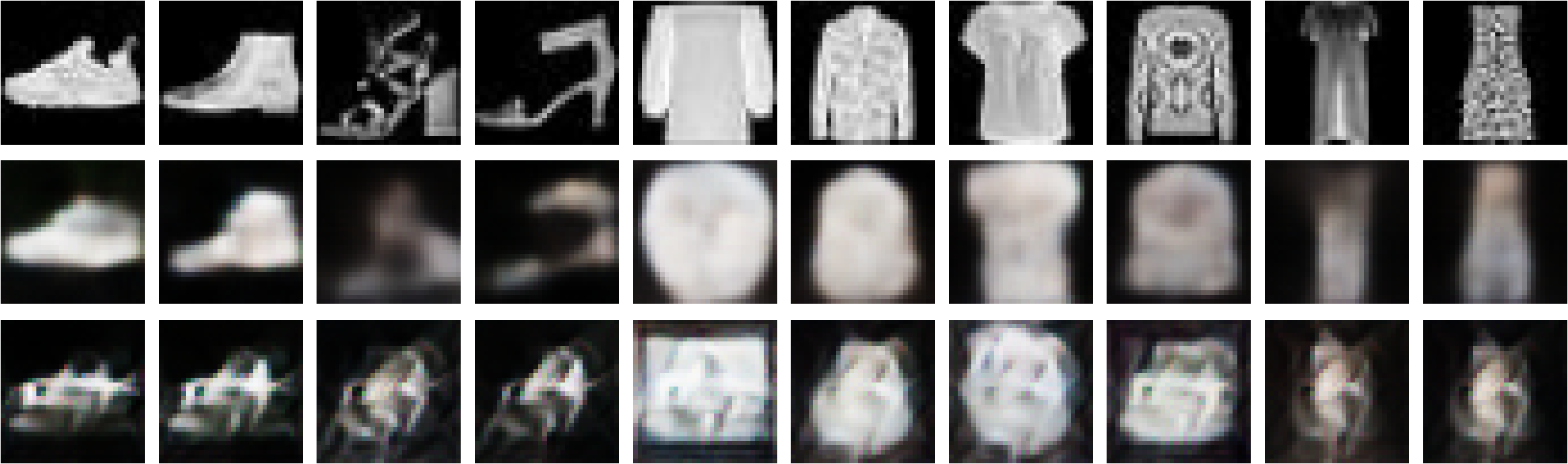}
		\caption{FashionMNIST (OOD)}
		\label{fig:fashionmnist-ood}
	\end{subfigure}
   \vspace{-0.2cm}
	\caption{Comparison of MSE and LPIPS loss: CIFAR10 (ID) and FashionMNIST (OOD). From top to bottom: Input, reconstruction (MSE), and reconstruction (LPIPS). ($L=32$)}
	\label{fig:mse-vs-perceptual}
    \vspace{-0.4cm}
\end{figure}

\Cref{tab:lpips-score} shows \gls{ood} detection performance using the \gls{lpips} score based on \gls{protodistvae} trained with either \gls{mse} or \gls{lpips} loss.
In contrast to using the \gls{mse} score which showed a generally lower performance (results not shown), the \gls{lpips} score can achieve good detection results, even when training with \gls{mse} reconstruction loss. However, using \gls{lpips} as reconstruction loss outperforms \gls{mse} loss. A special case is the ImageNet benchmark which is different due to image size and data diversity.
The reconstruction performance for MSE and LPIPS loss on the CIFAR10 benchmark is depicted in \Cref{fig:mse-vs-perceptual}. \gls{protodistvae} trained with \gls{mse} shows significant blur, regardless of ID or OOD samples. Training with LPIPS helps to preserve more semantic information and leads to differences when reconstructing OOD samples.

\begin{figure}[t]
	\centering
	\begin{subfigure}{0.4\textwidth}
		\centering
		\includegraphics[width=1\linewidth]{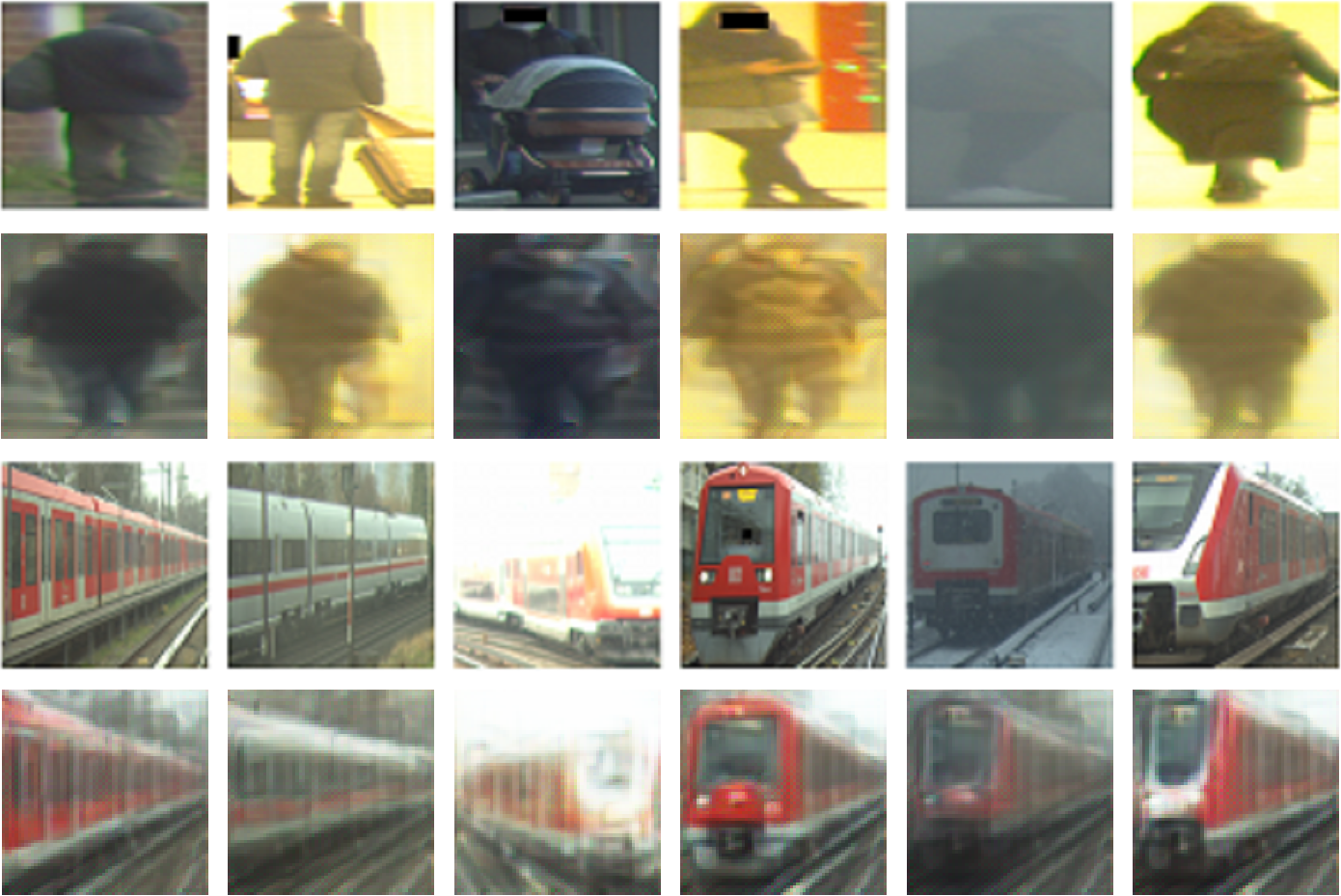}
		\caption{ID}
		\label{fig:rec-dbs-id}
	\end{subfigure}
        \hfill
	\begin{subfigure}{0.4\textwidth}
		\centering
		\includegraphics[width=1\linewidth]{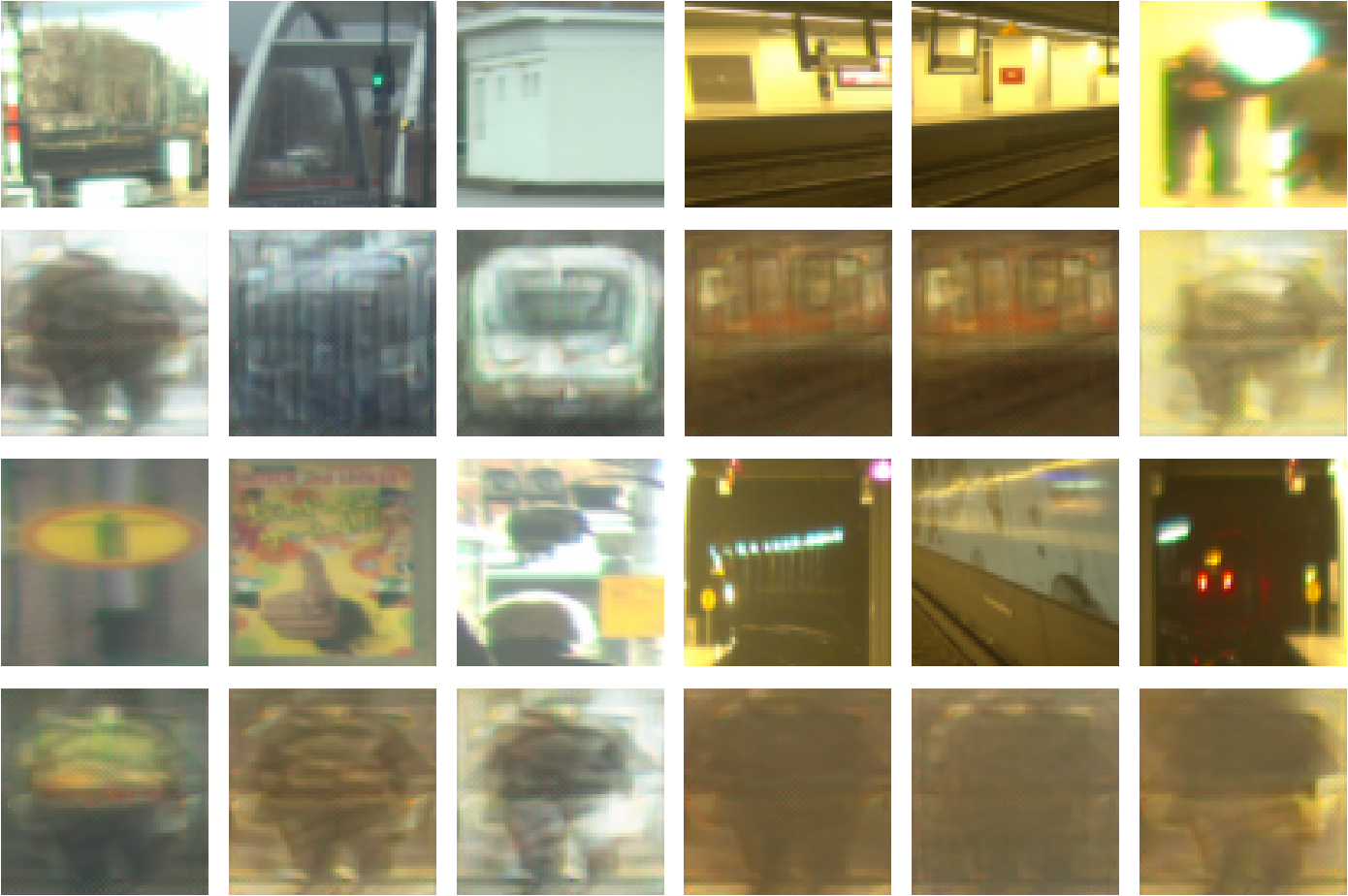}
		\caption{OOD}
		\label{fig:rec-dbs-ood}
	\end{subfigure}
    \vspace{-0.3cm}
	\caption{DBS samples and reconstructions: ID and OOD ($L\!=\!4096, w_\text{LPIPS}\!=\!100$)}
    \vspace{-0.4cm}
	\label{fig:rec-dbs-id-ood}
\end{figure}

\begin{figure}[t]
	\centering
	\begin{subfigure}{0.54\textwidth}
		\centering
		\includegraphics[width=1\linewidth]{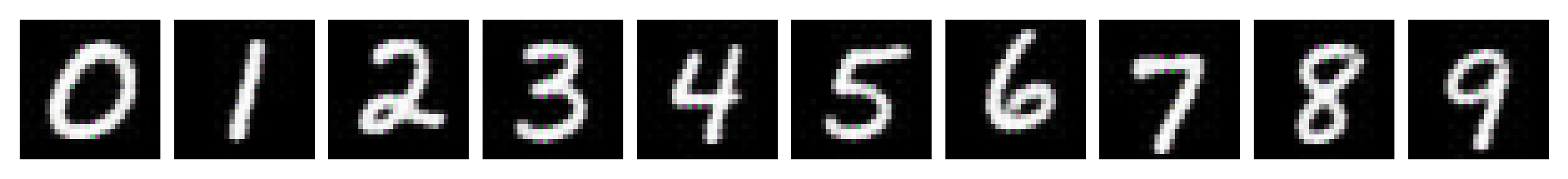}
            \includegraphics[width=1\linewidth]{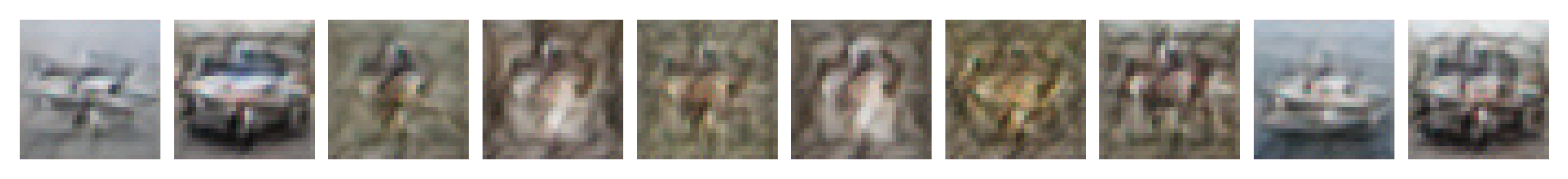}
            \caption{MNIST (top), CIFAR10 (bottom)}
		\label{fig:protorec-mnist-cifar10}
	\end{subfigure}
        \hfill
	\begin{subfigure}{0.39\textwidth}
		\centering
            \includegraphics[width=0.65\linewidth]{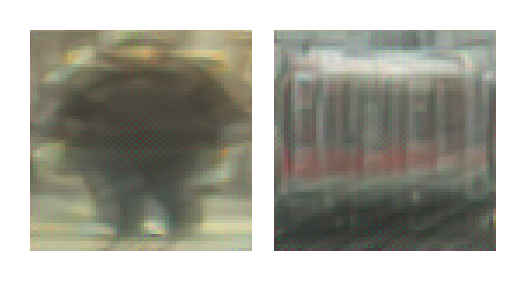}
            \caption{DBS ($L\!=\!4096, w_\text{LPIPS}\!=\!100$)}
		\label{fig:protorec-dbs}
	\end{subfigure}
        \vspace{-0.2cm}
	\caption{Prototype reconstructions of different benchmarks}
	\label{fig:protorec}
    \vspace{-0.4cm}
\end{figure}

\Cref{fig:rec-dbs-id-ood} displays reconstructions of the DBS dataset. 
\gls{protodistvae} appears to have learned the data distribution and can reconstruct \gls{id} better than \gls{ood} in most cases. It successfully distinguishes the class distributions of persons and trains and can show the features associated with a certain sample. For example, images of train stations and regular structures are often associated with trains, whereas background images are often reconstructed into person-like images.
The learned prototypes of \gls{protodistvae} can also be reconstructed.
As \Cref{fig:protorec} shows, prototypes can be better extracted from datasets with low-variance datasets like MNIST and the DBS dataset while for datasets with higher diversity like CIFAR10, prototypes are harder to extract and images are less expressive. Human observers can thus assess which properties the network extracted from the data and evaluate features associated across classes.

\section{Conclusion}

In this work, \gls{protodistvae} was applied to \gls{ood} detection. Its classification based on data density estimation, reconstruction, and prototype distances makes the approach robust and enables human insights into the classification and \gls{ood} detection process. The reconstruction of embeddings and prototypes enhances the model transparency w.r.t. properties learned from the data additionally.

The proposed method is extensively evaluated against state-of-the-art approaches using OpenOOD \cite{Yang2022} and real-world railway data.
Combining the estimation of the \gls{id} region with a perceptual reconstruction metric as well as integrating the proposed principle of enclosing restriction prove to be a good basis while 
the simple score fusion results in a gradual \gls{ood} score without imposing too many modeling assumptions.
As empirically shown for the real-world railway dataset, the model learns \acrlong{id} training data and can reliably exploit it for \gls{ood} detection. The reconstruction error is an additional cue to identify system failure. Future work could involve combining an enclosing restriction with other reconstruction-based approaches and extending the presented framework to the task of object detection.
Adversarial training may be used to increase the overall robustness.

\bibliographystyle{splncs04}
\bibliography{further/main}
\end{document}